\documentclass[10pt,twocolumn,letterpaper]{article}

\usepackage{wacv}                          

\usepackage{algorithm}
\usepackage{algorithmic}
\usepackage{booktabs}
\usepackage{multirow}
\usepackage{amssymb}
\usepackage{pifont}
\usepackage{xcolor}
\usepackage{enumitem}

\newcommand{\Asyn}{Q}

\newcommand{\indic}{\mathbb{1}}

\definecolor{okgreen}{rgb}{0.05,0.55,0.20}
\definecolor{badred}{rgb}{0.80,0.10,0.10}
\definecolor{zerogrey}{rgb}{0.55,0.55,0.55}
\newcommand{\cmark}{\ding{51}}
\newcommand{\xmark}{\ding{55}}
\newcommand{\ok}[1]{{\color{okgreen}#1~\cmark}}
\newcommand{\bad}[1]{{\color{badred}#1~\xmark}}
\newcommand{\zero}[1]{{\color{zerogrey}#1}}

\definecolor{wacvblue}{rgb}{0.21,0.49,0.74}
\usepackage[pagebackref,breaklinks,colorlinks,allcolors=wacvblue]{hyperref}

\def\confName{WACV}
\def\confYear{2027}

\title{Synthetic CT as a First Safeguard for Choosing Medical VLMs in Lung Cancer Screening}

\author{Fakrul Islam Tushar \qquad Geoffrey D. Rubin\\
{\tt\small fitushar@arizona.edu \qquad grubin@arizona.edu}\\
Department of Radiology and Imaging Sciences\\
University of Arizona}

\begin{document}
\maketitle
\begin{abstract}
Many medical vision--language models (VLMs) are publicly available, but a
site adopting one for a new imaging task rarely has the labeled real data
needed to compare them. Because synthetic CT has ground truth by
construction, a site could instead rank candidate models on synthetic scans,
if the ranking carries over to real data. We study when it does for lung CT,
using five public VLMs and three routine screening tasks, namely detecting a
nodule, locating its lobe, and estimating its size, on 85{,}021 synthetic and
26{,}085 real slices from seven datasets. For nodule presence and size, the
synthetic ranking matches the real one with a rank correlation of 0.98. For
lobe, the ranking is only moderately preserved, with a rank correlation of
0.80. We also propose TrialCouncil, a training-free method that uses
synthetic accuracy to select one model, combine several, or defer uncertain
cases. On real CT, it outperforms standard voting methods in every lobe and
size setting and outperforms the best synthetic model on lobe, while
selecting the best synthetic model works better for presence. Synthetic CT
can therefore serve as a first safeguard when choosing a medical VLM without
labeled real data.
\end{abstract}

\section{Introduction}\label{sec:intro}

\begin{figure*}[t]
  \centering
  \includegraphics[width=0.9\linewidth]{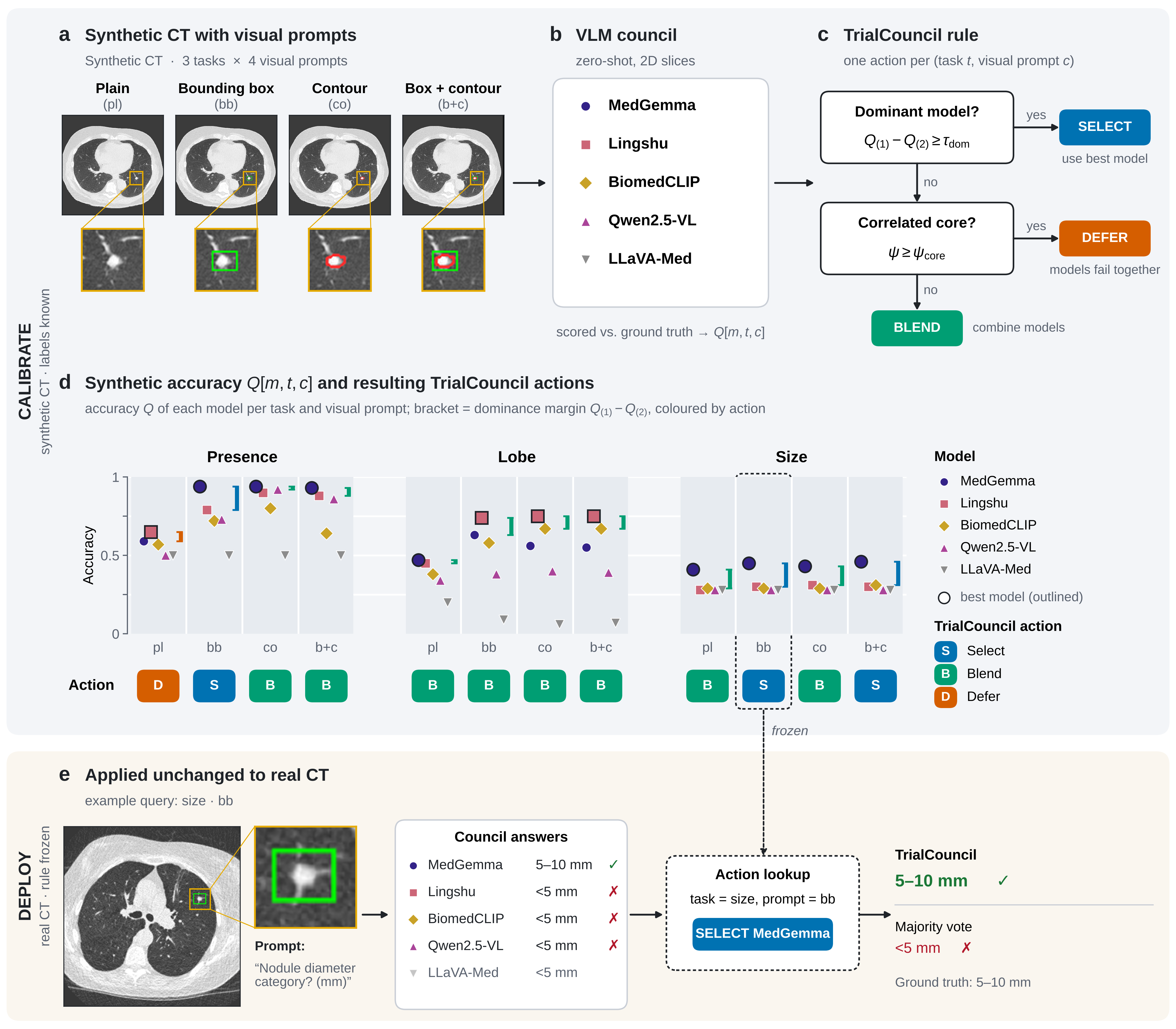}
  \caption{\textbf{How TrialCouncil works.} On synthetic CT, (a) each question is
  asked under four visual prompts, (b) five public VLMs answer and are scored
  against the known labels, and (c, d) TrialCouncil chooses one action per
  setting, selecting the leading model, blending with a weighted vote, or
  flagging low-confidence cases for deferral. (e) The chosen actions are
  applied unchanged to real CT, with no real labels and no retraining.}
  \label{fig:overview}
\end{figure*}

Lung cancer is the leading cause of cancer death in the United States,
accounting for about one in five cancer
deaths~\cite{siegel2025cancer,acs2026lungstats}. Low-dose CT screening has
been shown to be effective in reducing lung cancer
mortality~\cite{nlst2011reduced,dekoning2020nelson}. In screening, the
radiologist detects suspicious nodules, reports where each one is located,
measures its size, and assigns a category under the standardized Lung-RADS
reporting system~\cite{christensen2024lungrads}.

Medical vision--language models (VLMs) are an active area of research, and a
growing number of open-access models have shown promising results on
medical images~\cite{zhang2023biomedclip,li2023llava,sellergren2025medgemma,lingshu2025,hartsock2024vision}.
Notable performance discrepancies have also been reported, however, across
demographic groups~\cite{xia2024cares} and across prompt
wordings~\cite{ceballos2024open}.
Before any of these VLMs is adopted for a targeted task such as lung cancer
screening, it is essential to know its capabilities and limitations on the
target hospital or cohort, so that patients are safeguarded. Establishing
this normally requires a labeled local test set, and labeling CT is slow,
needs expert readers, and is often restricted by privacy rules.

Synthetic CT can be a way around this bottleneck. Simulation-based virtual
imaging trials and CT generated by deep generative models have been
introduced and applied to a range of
tasks~\cite{badano2018evaluation,tushar2025virtual,tushar2026itrialspace,guo2025maisi,tushar2025nodmaisi}.
We hypothesize that the ranking of models and the limitations observed on
synthetic data translate to real data, and can therefore help safeguard the
analysis of real data. We therefore ask two questions. \textbf{Q1.} Does the
ranking of VLMs measured on synthetic CT match their ranking on real CT?
\textbf{Q2.} Can a simple policy
fitted on synthetic data choose, combine, or abstain across models well on
real CT?

We study five open-access VLMs on three screening tasks (nodule presence, lobe, and size), each posed under four
visual prompts ranging from the plain slice to a bounding box plus contour,
using 85{,}021 synthetic and 26{,}085 real slices from seven public datasets
(\cref{fig:overview}). The study makes two contributions.
\begin{enumerate}[leftmargin=*, labelsep=0.5em, itemsep=0pt, topsep=3pt]
  \item \textbf{A synthetic-to-real ranking study.} We test whether the
  ranking of five open-access VLMs measured on synthetic CT matches their
  ranking on real CT, across three screening tasks and four visual prompts,
  and repeat the test with each real dataset's nodules removed from the
  synthetic data. Using synthetic data only, we also examine why the ranking
  carries over for some tasks but not others.
  \item \textbf{TrialCouncil.} We describe a training-free policy that uses
  synthetic accuracy to select one model, combine several, or flag
  low-confidence cases for deferral, and compare it with standard ensembles
  on real CT.
\end{enumerate}

\section{Related Work}\label{sec:related}

\textbf{Medical VLMs and their evaluation.} Reviews~\cite{hartsock2024vision}
and benchmarks~\cite{xia2024cares} of medical VLMs, and evaluations of
clinical language models~\cite{ceballos2024open}, measure accuracy on labeled real data, and deep
networks in general can rely on shortcuts~\cite{geirhos2020shortcut}. We
instead ask whether a site can choose among such models \emph{without}
labeled real data.

\textbf{Estimating performance without target labels.} When the deployment
data are unlabeled, performance can be estimated from in-distribution
accuracy~\cite{miller2021accuracy} or from model confidence on the unlabeled
target data~\cite{garg2022leveraging}, and both kinds of approach have been evaluated
on distribution-shift benchmarks such as WILDS~\cite{koh2021wilds}. Synthetic test data have also
been used to evaluate models on rare subgroups~\cite{vanbreugel2023synthetic}.
In medical imaging, virtual imaging trials use physics-based or learned
simulators to evaluate imaging systems and AI models under controlled
conditions~\cite{badano2018evaluation,abadi2020virtual,tushar2025virtual,tushar2026utility}.
These works mainly check image realism or whether simulated accuracy matches
real accuracy. We explored whether the \emph{ordering} of models carries over.

\textbf{Combining and selecting pretrained models.} Pretrained models can be
combined without retraining. Common approaches include majority voting and
simple combination rules such as the sum and product
rules~\cite{kittler1998combining}, recently also applied
to large language models~\cite{trad2024ensemble}. Another is dynamic
selection, which picks the classifier that is locally most
accurate~\cite{woods1997combination}. Related ideas include
self-consistency~\cite{wang2022self} and mixtures of
experts~\cite{shazeer2017outrageously}. Voting benefits from members making
different errors, although the link between this diversity and ensemble
accuracy is not straightforward~\cite{kuncheva2003measures}. Confidence-weighted methods
also need a reliable confidence score, which most generative VLMs do not
provide because they return only a text answer. Because TrialCouncil can also
flag cases for deferral, it relates to selective prediction, in which a
model abstains on uncertain
inputs~\cite{chow2003optimum,geifman2017selective,cheng2023regression} or
defers them to an expert~\cite{mozannar2020consistent}. These methods
usually fit their rules on labeled target data. TrialCouncil instead
estimates each model's accuracy, and its deferral rule, from labeled
synthetic data, and unlike test-time adaptation~\cite{wang2020tent} it never
changes model weights.

\section{Study Setup}\label{sec:setup}

\subsection{Tasks and visual prompts}\label{sec:tasks}

Each test case is a single axial CT slice, and we ask three questions about it.
\begin{itemize}\setlength\itemsep{0pt}
  \item \textbf{Presence} (2 classes) asks whether the slice contains a
  pulmonary nodule. Chance is $\pi_t=0.50$.
  \item \textbf{Lobe} (5 classes) asks which lobe contains the nodule (right
  upper, middle, or lower; left upper or lower). Chance is $\pi_t=0.20$.
  \item \textbf{Size} (4 classes) asks whether the nodule diameter is $<$5,
  5--10, 10--20, or $\geq$20\,mm. Chance is $\pi_t=0.25$.
\end{itemize}
We chose these tasks because they are routine but important steps in lung cancer screening, where the radiologist detects nodules and records their location and size in the radiology report. The tasks also differ in what they depend on. Presence and size are properties of the nodule, whereas lobe depends on where the nodule sits in the lung, so the same nodule placed in a different patient can belong to a different lobe.

We show each slice in four ways, which we call \emph{visual prompts}. The
slice is either plain (pl) or has a bounding box (bb), a contour (co), or
both (b+c) drawn around the nodule. Each task--prompt pair is one
\emph{setting}. The markers are drawn only when a nodule is present, so on
guided presence a model can do well just by saying ``yes'' whenever it sees
a marker. We therefore treat guided presence as checking agreement with a
marker and only plain presence as true detection, and we report them
separately.

\subsection{Models and inference}\label{sec:models}

We use five publicly released models without modification. One is a
contrastive model, BiomedCLIP~\cite{zhang2023biomedclip}, and four are
generative VLMs, namely LLaVA-Med~\cite{li2023llava},
MedGemma~\cite{sellergren2025medgemma}, Lingshu~\cite{lingshu2025}, and
Qwen2.5-VL~\cite{bai2025qwen25vl}. All were pretrained by their developers on
real data, and we do no training or fine-tuning; ``training-free'' in this
paper refers only to our own method.

The four generative models receive the same prompt per task, for example
\emph{``Does this CT slice contain a pulmonary nodule? Answer Yes or No.''},
and the lobe and size prompts list the five lobes and the four size ranges as
answer options. Decoding is greedy (deterministic, at most 64 new tokens)
without in-context examples, and a fixed rule-based parser maps each
free-text answer to a class; every output in our data parsed to a valid
class. BiomedCLIP instead picks the class whose text description (for example,
\emph{``A pulmonary nodule in the right upper lobe''}) is most similar to the
image. Prompts, parsers, and decoding are identical for synthetic and real
CT.

\subsection{Data}\label{sec:data}

\begin{table}[t]
  \centering
  \caption{\textbf{Data.} One slice per nodule through its center, with nodule-free
  slices from the same datasets as negatives. Lobe and size are scored on
  nodule slices only.}
  \label{tab:data}
  \footnotesize
  \setlength{\tabcolsep}{4pt}
  \begin{tabular}{@{}lrrr@{}}
    \toprule
    Dataset & Patients & Nodule slices & Negative slices \\
    \midrule
    DLCS24~\cite{wang2025duke}        & 1{,}605 & 2{,}473 & 2{,}473 \\
    IMDCT~\cite{zhao2025integrated}   & 2{,}032 & 2{,}032 & 1{,}943 \\
    LNDbv4~\cite{pedrosa2019lndb,ferreira2024lndbv4}     & 209     & 743     & 743 \\
    LUNA16~\cite{setio2017validation} & 599     & 1{,}179 & 1{,}179 \\
    LUNA25~\cite{peeters2025luna25}   & 2{,}120 & 6{,}156 & 6{,}156 \\
    LUNGx~\cite{armato2016lungx}      & 70      & 83      & 83 \\
    NSCLCR~\cite{aerts2014nsclc}      & 421     & 421     & 421 \\
    \midrule
    \textbf{Real total}               & \textbf{7{,}056} & \textbf{13{,}087} & \textbf{12{,}998} \\
    \textbf{Synthetic}~\cite{tushar2026itrialspace} & -- & \textbf{42{,}858} & \textbf{42{,}163} \\
    \bottomrule
  \end{tabular}
\end{table}

\begin{figure}[t]
  \centering
  \includegraphics[width=\linewidth]{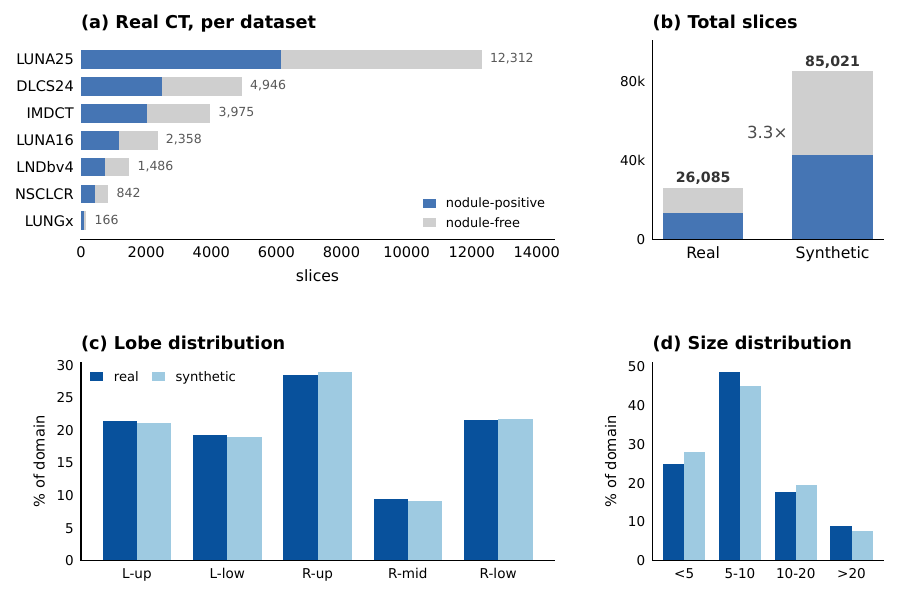}
  \caption{\textbf{Real and synthetic data.} (a) Nodule-positive and
  nodule-free slices in each real dataset. (b) Total slices, with the
  synthetic set about 3.3$\times$ larger than the real set. (c) Lobe and
  (d) size labels as a percentage of each domain.}
  \label{fig:dataset}
\end{figure}

\textbf{Real CT.} We pool seven public lung-CT datasets (\cref{tab:data}, \cref{fig:dataset}a),
giving 11{,}732 CT volumes from 7{,}056 patients with 13{,}087 annotated
nodules. Real data are used \emph{only} for evaluation.

\textbf{Synthetic CT.} We use the public iTrialSpace lung
dataset~\cite{tushar2026itrialspace} as released and did not generate it for
this work. It is built by profiling real nodules (the donors) from the same
public datasets, placing nodule masks with these properties at anatomically
plausible positions in patient scans (the hosts), and re-rendering each slice with a
ControlNet-conditioned generative model, so the presence, lobe, and size labels are known from the generation process. The dataset includes paired \emph{twins}, in which the same
nodule is placed back into its original scan and into a scan from a
different source dataset; we use them in \cref{sec:q1}. The synthetic
set is about 3.3$\times$ larger than the real set and has a similar
distribution of lobe and size labels (\cref{fig:dataset}b--d).

The Fr\'echet distance of this synthetic data has been reported
previously~\cite{tushar2026itrialspace}. Computed on 2.5D RadImageNet
ResNet-50 features, it falls within the range seen between pairs of real
datasets (median 1.57, IQR [1.05, 2.72]).

\subsection{Metrics and statistics}\label{sec:stats}

TrialCouncil selects models by accuracy (balanced accuracy for presence and accuracy on nodule-positive slices for lobe and size; \cref{sec:table}), whereas all results are reported as balanced accuracy for presence and macro-F1 for lobe and size, always per setting rather than pooled. Since a patient can contribute several
nodules, all 95\% confidence intervals (CIs) come from a patient-level
bootstrap that resamples whole patients with replacement within each dataset
(1{,}000 replicates for main results and 500 for method comparisons). Two methods are compared on the same resamples, and a
difference counts as a win or loss when its paired CI excludes zero and as a
tie otherwise.

\section{Method}\label{sec:method}

Everything in this section is computed on \emph{synthetic} CT only. We write
$m \in \{1,\dots,5\}$ for a model, $s=(t,c)$ for a setting with task $t$ and
visual prompt $c$, $a_m$ for the answer of model $m$ on a case, and $y$ for
the true label.

\begin{algorithm}[t]
\caption{TrialCouncil for one setting $s=(t,c)$}
\label{alg:gate}
\small
\begin{algorithmic}[1]
\REQUIRE synthetic accuracies $Q(\cdot,s)$; chance $\pi_t$;
$\tau_{\mathrm{dom}}{=}0.15$, $\psi_{\mathrm{core}}{=}0.5$; abstention cost $\lambda$
\STATE \textit{Calibration (synthetic CT only)}
\STATE compute weights $w_m$ (\cref{eq:weights}) and joint-miss rate $\psi$
\IF{$Q_{(1)} - Q_{(2)} \ge \tau_{\mathrm{dom}}$}
  \STATE action $\leftarrow$ \textsc{select} the top model $m^\star$
\ELSIF{$\psi \ge \psi_{\mathrm{core}}$}
  \STATE action $\leftarrow$ \textsc{defer}; fit $\kappa$ bins and $\theta^\star$ (\cref{eq:theta})
\ELSE
  \STATE action $\leftarrow$ \textsc{blend}
\ENDIF
\STATE \textit{Inference (real CT, answers $\{a_m\}$, no labels)}
\STATE $\hat{y} \leftarrow a_{m^\star}$ if \textsc{select}, else weighted vote
\STATE \textbf{if} \textsc{defer} and $\kappa < \theta^\star$ \textbf{then return} abstain \textbf{else return} $\hat{y}$
\end{algorithmic}
\end{algorithm}

\subsection{Synthetic accuracy table}\label{sec:table}

For every model and setting we measure the synthetic accuracy $Q(m,s)$. This
$5 \times 12$ table (\cref{fig:overview}d) sets the vote weights and decides
when one model is selected. On real CT, accuracy and macro-F1 pick the same
top model in 7 of 8 lobe and size settings (mean Spearman 0.96 between the
two rankings), so the choice of selection metric has little effect on the
results.

\begin{table*}[t]
  \centering
  \caption{\textbf{Real-CT results.} Balanced accuracy for presence and macro-F1 for
  lobe and size. In the top block, bold marks the real-best model and
  underline the synthetic-best model. In the bottom block, which uses no real
  labels, bold marks the best method and $^{+}$/$^{-}$ marks TrialCouncil as above
  or below the synthetic-best model (paired 95\% CI excluding zero). 95\% CIs are
  within $\pm$0.02.}
  \label{tab:main}
  \small
  \setlength{\tabcolsep}{4.2pt}
  \begin{tabular}{@{}lcccccccccccc@{}}
    \toprule
    & \multicolumn{4}{c}{Presence} & \multicolumn{4}{c}{Lobe} & \multicolumn{4}{c}{Size} \\
    \cmidrule(lr){2-5}\cmidrule(lr){6-9}\cmidrule(lr){10-13}
    & pl & bb & co & b+c & pl & bb & co & b+c & pl & bb & co & b+c \\
    \midrule
    BiomedCLIP~\cite{zhang2023biomedclip} & .553 & .664 & .708 & .626 & \textbf{.484} & \textbf{.739} & \textbf{.768} & \textbf{.757} & .160 & .190 & .149 & \textbf{.232}$^{\dagger}$ \\
    LLaVA-Med~\cite{li2023llava}  & .501 & .502 & .502 & .502 & .114 & .060 & .043 & .044 & .100 & .100 & .100 & .100 \\
    MedGemma~\cite{sellergren2025medgemma}   & .608 & \underline{\textbf{.958}} & \underline{\textbf{.956}} & \underline{\textbf{.963}} & \underline{.298} & .401 & .330 & .318 & \underline{\textbf{.308}} & \underline{\textbf{.207}} & \underline{\textbf{.254}} & \underline{.215} \\
    Lingshu~\cite{lingshu2025}    & \underline{\textbf{.653}} & .725 & .859 & .831 & .316 & \underline{.558} & \underline{.572} & \underline{.569} & .123 & .130 & .139 & .129 \\
    Qwen2.5-VL~\cite{bai2025qwen25vl} & .500 & .700 & .847 & .804 & .163 & .170 & .170 & .173 & .100 & .103 & .103 & .102 \\
    \midrule
    Plurality vote       & .599 & .847 & .931 & .915 & .326 & .495 & .406 & .398 & .100 & .102 & .101 & .116 \\
    Soft voting          & .598 & .846 & .929 & .914 & .297 & .413 & .332 & .330 & .102 & .107 & .104 & .116 \\
    Confidence-weighted  & .599 & .847 & .931 & .915 & .297 & .413 & .332 & .330 & .101 & .106 & .104 & .116 \\
    Synthetic-best model & \textbf{.653} & \textbf{.958} & \textbf{.956} & \textbf{.963} & .298 & .558 & .572 & .569 & \textbf{.308} & \textbf{.207} & \textbf{.254} & \textbf{.215} \\
    \textbf{TrialCouncil (ours)} & .599$^{-}$ & \textbf{.958} & .924$^{-}$ & .906$^{-}$ & \textbf{.351}$^{+}$ & \textbf{.619}$^{+}$ & \textbf{.624}$^{+}$ & \textbf{.631}$^{+}$ & \textbf{.308} & \textbf{.207} & \textbf{.254} & \textbf{.215} \\
    \bottomrule
  \end{tabular}\\[2pt]
  {\footnotesize $^{\dagger}$Under accuracy, the metric used for selection,
  MedGemma is best in this setting too.}
\end{table*}

\subsection{TrialCouncil}\label{sec:council}

TrialCouncil turns the synthetic accuracy table into one of three actions for
each setting (\cref{alg:gate}). It selects a single model when that model's
synthetic accuracy is at least 0.15 above every other model's. Otherwise it takes an accuracy-weighted vote, and it
adds a deferral step when the useful models often fail on the same cases.

\textbf{Model weights.} Each model gets a weight proportional to how far its
accuracy is above chance,
\begin{equation}
  w_m(s) \;\propto\; \max\!\Big(0,\;
  \frac{\Asyn(m,s)-\pi_t}{1-\pi_t}\Big),
  \qquad \textstyle\sum_m w_m(s) = 1,
  \label{eq:weights}
\end{equation}
so a model at or below chance has no influence. LLaVA-Med, which answers
``present'' on almost every presence slice, gets zero weight on presence and
lobe and a small weight on size.

\textbf{Select.} Let $Q_{(1)} \ge Q_{(2)}$ be the two highest synthetic
accuracies in a setting. If $Q_{(1)} - Q_{(2)} \ge \tau_{\mathrm{dom}} =
0.15$, TrialCouncil returns the answer of the top model. This is model
selection by synthetic accuracy.

\textbf{Blend.} Otherwise TrialCouncil takes the weighted vote
$\hat{y} = \arg\max_{y} \sum_m w_m(s)\,\indic[a_m = y]$. If one model holds
more than half of the total weight, the vote always follows that model, and
blending becomes selection.

\textbf{Defer.} No vote can fix a case on which every useful model is wrong.
We therefore measure the \emph{joint-miss rate} $\psi(s)$ on synthetic
nodule-positive slices. It is the fraction of cases on which all models
except the near-chance LLaVA-Med are wrong. We compare it with the rate
$\prod_m q_m(s)$ expected if errors were independent, where $q_m$ is the miss
rate of model $m$. The ratio $r(s) = \psi(s) / \prod_m q_m(s)$ is above 1
when the models fail together. If no model is selected and $\psi(s) \ge
\psi_{\mathrm{core}} = 0.5$, TrialCouncil still votes, but it also gives
each answer a confidence $\kappa$ so that low-confidence cases can go to a
human reader.

We build $\kappa$ on synthetic data. Each case is scored by the weighted vote
share of the winning answer, $v = \max_y \sum_m w_m \indic[a_m=y]$. Cases are
grouped into five equal-size bins of $v$ (separately for each predicted class
in presence), and $\kappa$ is the accuracy of the vote within each bin. On
real CT, each case's $v$ is mapped to the $\kappa$ of its bin, and
TrialCouncil answers only if $\kappa \ge \theta^\star$, where
\begin{equation}
  \theta^\star = \arg\min_{\theta}\;
  \big[\mathrm{risk}(\theta) + \lambda\,(1-\mathrm{coverage}(\theta))\big].
  \label{eq:theta}
\end{equation}
Coverage is the fraction of cases answered, and risk is the error rate on
those cases. The cost of abstaining, $\lambda$, is chosen by the deploying
site. We use 0.5 by default, and \cref{sec:abstain} shows how other values
change coverage.

All thresholds ($\tau_{\mathrm{dom}}$, $\psi_{\mathrm{core}}$, $\lambda$)
were fixed before any real evaluation. No setting in our data meets both the
select and the defer condition. On synthetic CT the rule chooses
\textsc{select} in 3 settings (presence-bb, size-bb, size-b+c),
\textsc{defer} in 1 (presence-pl), and \textsc{blend} in the other 8
(\cref{fig:overview}).

\subsection{Evaluation protocol and baselines}\label{sec:protocol}

\textbf{Synthetic-to-real transfer.} TrialCouncil's weights, actions,
$\kappa$ bins, and abstention threshold $\theta^\star$ are fitted on
synthetic CT and then applied to real CT unchanged. To measure what is lost by fitting on
synthetic rather than real data, we also fit the same rule on a real training
split and test both versions on a held-out real split. The \emph{transfer
gap} is the score of the real-fitted rule minus that of the synthetic-fitted
rule on this test split.

\textbf{Shared-source check.} The synthetic nodules are generated from masks
of nodules in the same public datasets we test on. To check that this overlap
does not explain the transfer, we hold out one real dataset at a time, remove
the synthetic nodules generated from it, and recompute synthetic accuracy. We
then check whether the synthetic-best model is still the real-best model on
the held-out dataset (\emph{leave-source-out}).

\textbf{Baselines.} We compare TrialCouncil with \emph{plurality} (majority)
voting, with \emph{soft voting}, which averages class scores, and with
\emph{confidence-weighted} voting, which weights each vote by the model's own
confidence. Because four of the five models output only a label, soft and
confidence-weighted voting give nearly the same results, and on presence and
size they also nearly match plurality. The main baseline is the
\emph{synthetic-best model}, which uses the model with the highest synthetic
accuracy in each setting. As an upper reference we also report the
\emph{real-best model}, which needs real labels. For abstention, we
compare ranking cases by $\kappa$ with ranking them by model confidence, by
vote entropy, and at random.

\raggedbottom
\section{Results}\label{sec:results}

\begin{figure*}[t]
  \centering
  \includegraphics[width=0.85\linewidth]{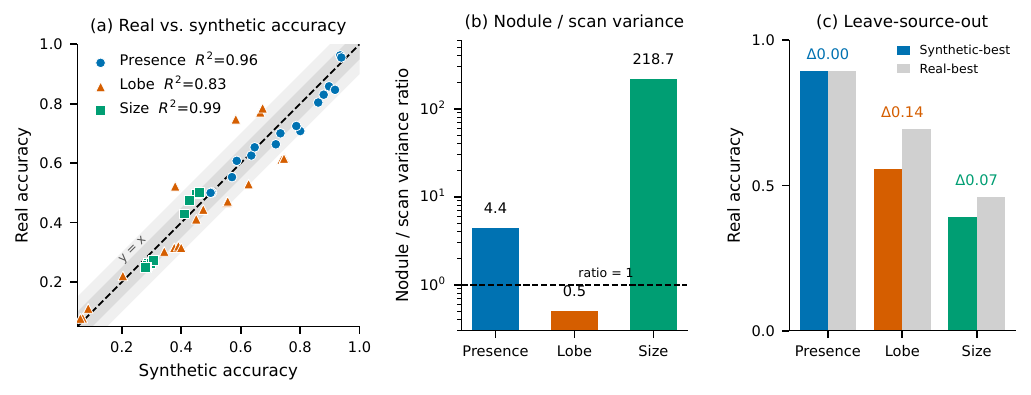}
  \caption{\textbf{Which synthetic results carry over to real CT.} (a) Real versus
  synthetic accuracy, one point per model and prompt. (b) Nodule/scan
  variance ratio from the synthetic twins. Lobe is the only task below 1.
  (c) Leave-source-out real accuracy of the synthetic-best model (color) and
  the real-best model (grey).}
  \label{fig:transfer}
\end{figure*}

\subsection{Does the model ranking transfer (Q1)?}\label{sec:q1}

\textbf{Presence and size.} The model that scores best on
synthetic CT is also the best on real CT in all four presence and all four
size settings (\cref{tab:main}). Real accuracy rises linearly with
synthetic accuracy ($R^2 = 0.96$ for presence and $0.99$ for size;
\cref{fig:transfer}a), and the mean Spearman correlation between the two
rankings is 0.98 for both. On the plain slice, which measures unaided
detection, Lingshu is best on both synthetic (0.65) and real CT (0.653). For
size, only MedGemma and BiomedCLIP score well above the 0.10 of a trivial
predictor, so this result mainly shows which of the two leads.

\begin{figure*}[t]
  \centering
  \scriptsize
  \setlength{\tabcolsep}{3pt}
  \begin{tabular}{@{}ccccc@{}}
    \textbf{(a) \textsc{select} (size, bb)} &
    \textbf{(b) \textsc{blend} (lobe, pl)} &
    \textbf{(c) \textsc{defer} (presence, pl)} &
    \textbf{(d) Failure (lobe, co)} &
    \textbf{(e) Failure (size, pl)} \\
    \includegraphics[width=0.165\linewidth]{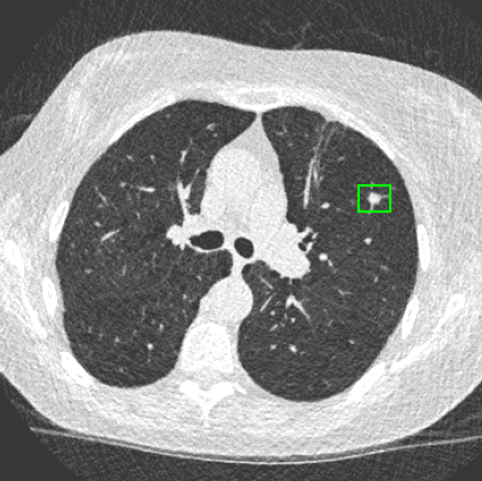} &
    \includegraphics[width=0.165\linewidth]{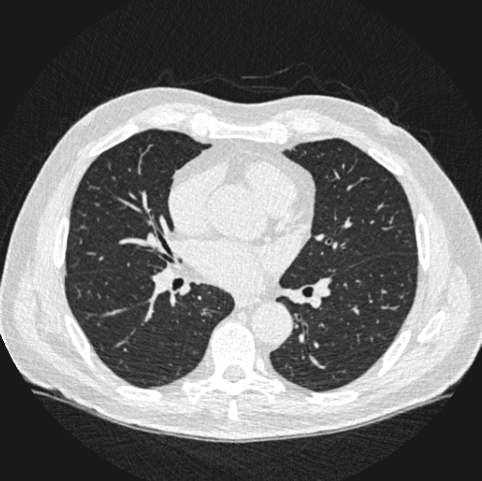} &
    \includegraphics[width=0.165\linewidth]{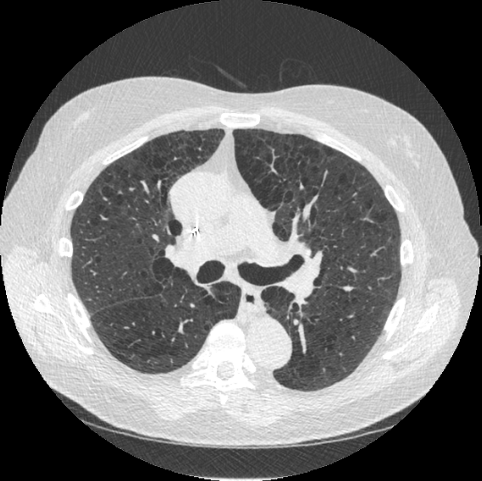} &
    \includegraphics[width=0.165\linewidth]{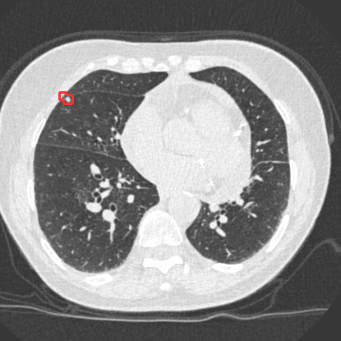} &
    \includegraphics[width=0.165\linewidth]{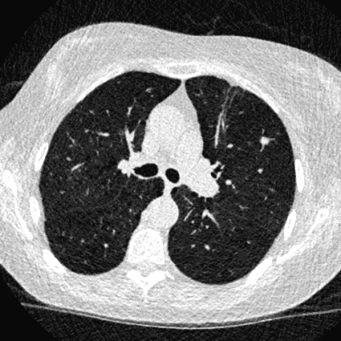} \\[2pt]
    \begin{minipage}[t]{0.185\linewidth}\raggedright
      BiomedCLIP: \bad{$<$5\,mm}\\ MedGemma: \ok{5--10\,mm}\\
      Lingshu: \bad{$<$5\,mm}\\ Qwen2.5-VL: \bad{$<$5\,mm}\\
      \zero{LLaVA-Med: $<$5\,mm}\\[2pt]
      \textbf{TC selects MedGemma:} \ok{5--10\,mm}\\ Truth: 5--10\,mm
    \end{minipage} &
    \begin{minipage}[t]{0.185\linewidth}\raggedright
      BiomedCLIP: \ok{R lower}\\ MedGemma: \bad{R upper}\\
      Lingshu: \bad{R middle}\\ Qwen2.5-VL: \ok{R lower}\\
      \zero{LLaVA-Med: L upper}\\[2pt]
      \textbf{TC weighted vote:} \ok{R lower}\\ Truth: R lower
    \end{minipage} &
    \begin{minipage}[t]{0.185\linewidth}\raggedright
      BiomedCLIP: \bad{absent}\\ MedGemma: \bad{absent}\\
      Lingshu: \bad{absent}\\ Qwen2.5-VL: \bad{absent}\\
      \zero{LLaVA-Med: present}\\[2pt]
      \textbf{TC} flags it for deferral (low $\kappa$)\\ Truth: present
    \end{minipage} &
    \begin{minipage}[t]{0.185\linewidth}\raggedright
      BiomedCLIP: \ok{R middle}\\ MedGemma: \bad{R upper}\\
      Lingshu: \bad{R upper}\\ Qwen2.5-VL: \bad{R upper}\\
      \zero{LLaVA-Med: L upper}\\[2pt]
      \textbf{TC weighted vote:} \bad{R upper}\\ Truth: R middle
    \end{minipage} &
    \begin{minipage}[t]{0.185\linewidth}\raggedright
      BiomedCLIP: \bad{$<$5\,mm}\\ MedGemma: \bad{$<$5\,mm}\\
      Lingshu: \bad{$<$5\,mm}\\ Qwen2.5-VL: \bad{$<$5\,mm}\\
      \zero{LLaVA-Med: $<$5\,mm}\\[2pt]
      \textbf{TC weighted vote:} \bad{$<$5\,mm}\\ Truth: 5--10\,mm
    \end{minipage}
  \end{tabular}
  \caption{\textbf{Real-CT examples.} (a--c) One case for each action. (d) A lobe
  failure where the only correct model, BiomedCLIP, is under-weighted. (e) A
  case where every model is wrong. LLaVA-Med, with zero or small weight, is
  shown in grey.}
  \label{fig:qual}
\end{figure*}

\textbf{Lobe.} The order of models is moderately preserved
($R^2 = 0.83$, Spearman 0.80), but the top model changes. Lingshu or MedGemma
ranks first on synthetic CT, while BiomedCLIP is best on real CT (macro-F1 up
to 0.77), so the synthetic-best model is never the real-best one.

\textbf{Why lobe differs.} The synthetic twins, in which the same nodule is
placed into scans from different source datasets, let us separate the effect
of the nodule from that of the scan around it. We compute model accuracy for
each source dataset of the nodule and for each source dataset of the scan,
and compare the variance across the two. For presence and size, accuracy
varies more with the nodule (variance ratio 4.4 and 218.7). For lobe the
ratio is 0.5, so accuracy depends more on the scan (\cref{fig:transfer}b,
\cref{tab:diagnostic}). This fits the lobe result, since the lobe of a nodule
depends on where it sits in the patient's lung. The size ratio is large
mainly because size accuracy barely changes across scan sources, so only its
direction is meaningful.

\textbf{Shared data sources.} When the synthetic nodules generated from each
real dataset are removed in turn, the synthetic-best model still matches the
real-best model on the held-out dataset in 26 of 28 presence cases, 19 of 28
size cases, and 0 of 28 lobe cases (7 datasets $\times$ 4 prompts). The agreement for presence and size is therefore not solely explained by
overlap in the source nodules.
The real accuracy lost by using the synthetic choice is 0.002
for presence, 0.070 for size, and 0.138 for lobe (\cref{fig:transfer}c). For
size, most of this loss comes from two small datasets, LNDbv4 and NSCLCR.

\textbf{Fitting on synthetic versus real labels.} Fitted on real labels
instead, TrialCouncil makes the same choices for presence and size, and the
transfer gap is 0.000 on all seven datasets. For lobe, real labels help on
every dataset (mean gap $+0.07$, range 0.018--0.095). A zero gap means the two
versions score the same, not that the score is high.

\begin{table}[t]
  \centering
  \caption{\textbf{Synthetic-to-real transfer by task.} The top block uses
  synthetic twins only and the bottom block real CT. Top-1 counts settings
  where the synthetic-best model is also the real-best model, and LSO is
  leave-source-out. $^{\ddagger}$By accuracy; by macro-F1 one size setting
  differs by 0.017.}
  \label{tab:diagnostic}
  \footnotesize
  \setlength{\tabcolsep}{3pt}
  \begin{tabular}{@{}lccc@{}}
    \toprule
    & Presence & Size & Lobe \\
    \midrule
    \multicolumn{4}{@{}l}{\emph{Synthetic twins only}} \\
    Accuracy range across nodule sources & 0.133 & 0.238 & 0.056 \\
    Accuracy range across scan sources & 0.061 & 0.017 & 0.079 \\
    Nodule/scan variance ratio & 4.4 & 218.7 & \textbf{0.5} \\
    \midrule
    \multicolumn{4}{@{}l}{\emph{Real CT}} \\
    $R^2$ (real vs.\ synthetic) & 0.96 & 0.99 & \textbf{0.83} \\
    Spearman (model ranks) & 0.98 & 0.98 & \textbf{0.80} \\
    Top-1, pooled real data$^{\ddagger}$ & 4/4 & 4/4 & \textbf{0/4} \\
    Top-1, LSO & 26/28 & 19/28 & \textbf{0/28} \\
    Accuracy lost, LSO & 0.002 & 0.070 & \textbf{0.138} \\
    Transfer gap (mean, 7 datasets) & 0.000 & 0.000 & \textbf{+0.07} \\
    \bottomrule
  \end{tabular}
\end{table}

\subsection{How well does TrialCouncil do on real CT (Q2)?}\label{sec:q3}

\Cref{tab:main} compares TrialCouncil with all baselines in each setting. The
most informative comparison is with the synthetic-best model, the simplest
strategy that needs no real labels. TrialCouncil outperforms all three
voting methods in every lobe and size setting.

\textbf{Size.} TrialCouncil returns MedGemma's answer in all four size
settings, either by selecting it (two settings) or because MedGemma holds
more than half of the vote weight. It therefore matches the
synthetic-best model and roughly triples the score of plurality voting (0.308
vs.\ 0.100 on the plain slice), in which the weaker models outvote MedGemma.

\textbf{Presence.} Here, selecting one model works better than blending.
With a bounding box, TrialCouncil selects MedGemma and matches the best model
(0.958). With a contour or both overlays, no model leads by 0.15 on synthetic
CT, so TrialCouncil blends and trails the synthetic-best model by 0.03--0.06.
On the plain slice, the \textsc{defer} setting, it equals plurality (0.599)
when it answers every case, so its value there lies in the confidence
$\kappa$ (\cref{sec:abstain}).

\textbf{Lobe.} TrialCouncil beats the synthetic-best model in all four lobe
settings (by $+0.05$ to $+0.06$) and plurality by up to
$+0.23$, yet it stays 0.12--0.14 below the real-best model (0.624 vs.\
0.768 with a contour; \cref{fig:qual}d). The real-best lobe model, BiomedCLIP,
ranks only second or third on synthetic CT and so receives too little weight.
The weighted vote cannot recover BiomedCLIP, and the size of the remaining
gap could not be known without real labels.

\subsection{Shared failures and abstention}\label{sec:abstain}

\textbf{The models fail together.} On real plain-slice nodules, all four
models other than LLaVA-Med miss the nodule in 50.5\% of cases, while
independent errors would predict 38.4\% ($r = 1.31$). For lobe, all four are
wrong on 23.3\% of cases versus 10.9\% expected ($r = 2.14$). Visual prompts
reduce the joint-miss rate for presence to 3--5\%, but the remaining errors
are still correlated ($r = 3.7$--$15.7$). No vote can recover a case that
every useful model gets wrong, so such cases can only be passed to a reader.
On synthetic CT the defer criterion ($\psi \ge 0.5$) is met only in plain
presence, which is why it is the \textsc{defer} setting.

\textbf{Confidence for abstention.} On unaided presence, ranking cases by
$\kappa$ and answering only the more confident half reduces the error from
0.40 to 0.25. By the area under the risk--coverage
curve (AURC), $\kappa$ beats model confidence by 0.057 [0.054, 0.060] on the
plain slice, and it beats confidence, entropy, and random ordering on all
four presence prompts. It gives no advantage on lobe and is worse than random
on guided size, so we report abstention as a presence result only.

\textbf{Choosing the operating point.} How much to abstain is a cost decision
for the site. With the default cost $\lambda = 0.5$ (\cref{eq:theta}), the
rule never abstains on unaided presence, whereas it abstains on 5\% of cases
at $\lambda = 0.3$ and on 26--27\% at $\lambda \le 0.2$.

\raggedbottom
\section{Discussion and Conclusion}\label{sec:discussion}

Our results show that model rankings on synthetic CT were more consistent
with those on real CT for nodule presence and size than for nodule lobe. Combining models helped
mainly for lobe, where no single model led on synthetic CT, while for presence
selecting the best synthetic model worked better than blending.

This study has several limitations. Our exploration is limited to synthetic
CT from a single deep generative model, although this model has shown better
generative performance than comparable
models~\cite{zhao2026maisiv2,tushar2025nodmaisi,tushar2026itrialspace}. A future study will repeat this
analysis with other generators. This study targeted three tasks, and a future study
will expand to more, such as nodule texture, margin, or distance to the
pleura.
We also did not compare with accuracy estimators that use unlabeled real
data~\cite{garg2022leveraging,baek2022agreement}. Lobe and size are read from
a single slice, whereas 3D input would be closer to clinical reading.
Since TrialCouncil only chooses among existing models, future work will
also explore fine-tuning these models on synthetic labels.

In conclusion, synthetic CT can serve as a first safeguard when choosing a
medical VLM without labeled real data, and it is a promising practice for
evaluating medical AI before deployment.

\section*{Acknowledgments}
This work was supported by startup funding from the Department of Radiology
and Imaging Sciences at the University of Arizona.

{
    \small
    \bibliographystyle{ieeenat_fullname}
    \bibliography{main}
}

\end{document}